\documentclass[runningheads]{llncs}

\usepackage{amsmath}
\usepackage{amsfonts}
\usepackage{amssymb}
\usepackage{graphicx}
\usepackage[TABBOTCAP]{subfigure}
\usepackage{color,graphicx,epic}

\usepackage{url}
\urldef{\mailsa}\path|{francois-xavier.dupe,luc.brun}@greyc.ensicaen.fr|
\newcommand{\keywords}[1]{\par\addvspace\baselineskip
\noindent\keywordname\enspace\ignorespaces#1}

\newcommand{\norm}[1]{\left\| #1 \right\|}
\newcommand{\abs}[1]{\left\lvert #1 \right\rvert}
\newcommand{\Kpath}{\mbox{$K_{path}$}}
\newcommand{\Kclassic}{\mbox{$K_{classic}$}}
\newcommand{\Kedit}{\mbox{$K_{edit}$}}

\newcommand{\Kchange}{\mbox{$K_{change}$}}
\newcommand{\KchangeC}{\mbox{$K_{change,classic}$}}

\newcommand{\Kmax}{\mbox{$K_{max}$}}
\newcommand{\KmaxC}{\mbox{$K_{max,classic}$}}

\newcommand{\Kmatching}{\mbox{$K_{matching}$}}
\newcommand{\KmatchingC}{\mbox{$K_{matching,classic}$}}

\newcommand{\Knew}{\mbox{$K_{new}$}}

\graphicspath{{images/}{images/hands/}{images/tools/}{images/dudes/}%
{images/others/}}

\begin{document}

\mainmatter  

\title{Hierarchical bag of paths for kernel based shape classification}


%
%
\author{Fran\c{c}ois-Xavier Dup\'e\thanks{This work is performed in close  collaboration with
    the laboratory Cyc\'eron and is supported by  the CNRS and the r\'egion Basse-Normandie.}
  \and Luc Brun
}
%

\institute{GREYC UMR CNRS 6072,\\
  ENSICAEN-Universit\'e de Caen Basse-Normandie,\\
  14050 Caen France,\\
  \mailsa
}

%
%

\maketitle

\begin{abstract}
  Graph kernels methods are based on an implicit embedding of graphs
  within a vector space of large dimension. This implicit embedding
  allows to apply to graphs methods which where until recently solely
  reserved to numerical data.  Within the shape classification
  framework, graphs are often produced by a skeletonization step which
  is sensitive to noise. We propose in this paper to integrate the
  robustness to structural noise by using a kernel based on a bag of
  path where each path is associated to a hierarchy encoding
  successive simplifications of the path.  Several experiments prove
  the robustness and the flexibility of our approach compared to
  alternative shape classification methods.
\end{abstract}

\section{Introduction}
\label{sec:intro}

The skeleton of a $2D$ shape is defined as the location of the
singularities of the signed distance function to the border of the
shape. This structure has several interesting properties: it is thin,
homotopic to the shape, invariant under rigid transformations of the
plane and most importantly it has a natural interpretation as a graph.
The representation of a shape by a skeletal (or shock) graph has
become popular owing the good properties of this representation
inherited from the properties of the skeleton.  However, beside all
this good properties, the skeletonization is not continuous and small
perturbations of the boundary insert structural noise within the graph
encoding the shape.

Several graph based methods have been proposed to compute a distance
between shapes robust to such a structural noise. Sharvit et
al.~\cite{sharvit-98} propose a graph matching method based on a
graduated assignment algorithm.  Siddiqi~\cite{sidiqqi-99} proposes to
transform the shock graph into a tree and then applies a tree matching
algorithm.  Pellilo~\cite{pelillo-99} uses the same tree
representation but transforms the tree matching problem into a maximal
clique problem within a specific association graph.

All the above graph methods operate directly on the space of graphs
which contains almost no mathematical structure. This lack of
mathematical structure forbids the use of basic statistical tools such
as the mean or the variance. Graph kernels provide an elegant solution
to this problem.  Using appropriate kernels, graphs can be mapped
either explicitly or implicitly into a vector space whose dot product
corresponds to the kernel function.  All the ``natural'' operations on
a set of graphs which were not defined in the original graph space are
now possible into this transformed vector space. In particular, graph
kernels may be combined with the kernelised version of robust
classification algorithms such as the Support Vector Machine (SVM).

A Graph kernel used within the shape representation framework should
take into account the structural noise induced by the skeletonization
process. Bunke~\cite{bunke-07} proposes to combine edit distance and
graph kernels by using a set of $n$ prototype graphs
$\{g_1,\dots,g_n\}$. Given a graph edit distance $d(.,.)$, Bunke
associates to each graph $g$ the vector $
\phi(g)=(d(g,g_1)\dots,d(g,g_n))$.  The kernel $k(g_1,g_2)$ between the
two graphs $g_1$ and $g_2$ is then defined as the dot product
$<\phi(g_1),\phi(g_2)>$.

Neuhaus~\cite{Neuhaus-06} proposes a similar idea by defining for a
prototype graph $g_0$, the kernel: \(
k_{g_0}(g,g')=\frac{1}{2}\left(d^2(g,g_0)+d^2(g_0,g')-d^2(g,g')\right)
\), where $d(.,.)$ denotes the graph edit distance.  Several graph
prototypes may be incorporated by summing or multiplying such kernels.
Using both Neuhaus~\cite{Neuhaus-06} and Bunke~\cite{bunke-07}
kernels two close graphs should have close edit distance to the
different graph prototypes. The metric induced by such graph kernels
is thus relative both to the weights used to define the edit distance
and to the graph prototypes. This explicit use of prototype graphs may
appear as artificial in some application.  Moreover, the definite
positive property of these kernels may not in general be guaranteed.

Suard~\cite{suard-07} proposes to use the notion of bag of paths of finite length for
shape matching.  This method associates to each graph all its paths whose length is lower
than a given threshold.  The basic idea of this approach is that two close shapes should
share a large amount of paths. A kernel between these sets should thus reflect this
proximity. However, small perturbations may drastically reduce the number of common paths
between two shapes (Section~\ref{subsec:kerhier}). Moreover, the straightforward
definition of a kernel between set of paths does not lead to a definite positive kernel
(Section~\ref{subsec:kernmax}).

This paper proposes a new definite positive kernel between set of
paths which takes into account the structural noise induced by the
skeletonization process.  We first present in
Section~\ref{sec:kernBag} the bag of paths approach for shape
similarity. Our contributions to this field are then presented in
Section~\ref{sec:kernhier}. The effectiveness of our method is
demonstrated through experiments in Section~\ref{sec:Experiments}.



\section{Kernels on bag of paths}
\label{sec:kernBag}

Let us consider a graph $G=(V,E)$ where $V$ denotes the set of
vertices and $E\subset V\times V$ the set of edges.  As mentioned in
Section~\ref{sec:intro} a bag of paths $P$ of length $s$ associated to
$G$ contains all the paths of $G$ of length lower than $s$. We denote
by $\abs{P}$ the number of paths inside $P$.  Let us denote by
$K_{path}$ a generic kernel between paths. Given two graphs $G_1$ and
$G_2$ and two paths $h_1\in P_1$ and $h_2\in P_2$ of respectively
$G_1$ and $G_2$, $K_{path}(h_1,h_2)$ may be interpreted as a measure
of similarity between $h_1$ and $h_2$ and thus as a local measure of
similarity between these two graphs. The aim of a kernel between bags
of paths consits to agregate all these local measures between pairs of
paths into a global similarity measure between the two graphs.

\subsection{The max kernel}
\label{subsec:kernmax}
This first method, proposed by Suard~\cite{suard-07}, uses the kernel
\Kpath{} as a measure of similarity and computes for each path $h_1\in
P_1$ the similarity with its closest path in $P_2$($\max_{h_j\in P_2}
\nobreakspace\Kpath(h_1,h_j)$). A first global measure of similarity
between $P_1$ and $P_2$ is then defined as:

\begin{equation}
  \label{eq:1}
  \hat{K}_{max}(G_1,G_2) = \hat{K}_{max}(P_1,P_2) = \frac{1}{\abs{P_1}} \sum_{h_i\in P_1}
  \max_{h_j\in P_2} K_{path}(h_i,h_j).
\end{equation}

The function $\hat{K}_{max}(G_1,G_2)$ is however not symmetric
according to $G_1$ and $G_2$. Suard obtains a symmetric function
interpreted as a graph kernel by taking the mean of
$\hat{K}_{max}(G_1,G_2)$ and $\hat{K}_{max}(G_2,G_1)$:
\begin{equation}
  \label{eq:6}
  \Kmax(G_1,G_2) = \tfrac{1}{2} \left[ \hat{K}_{max}(G_1,G_2) +  \hat{K}_{max}(G_2,G_1) \right].
\end{equation}

This kernel is not positive definite in general. However as shown by
Haasdonk \cite{Haasdonk2005}, SVM with indefinite kernels have in some
cases a geometrical interpretation as the maximization of distances
between convex hulls.  Moreover, experiments
(section~\ref{sec:Experiments}, and~\cite{suard-07}) show that this
kernel usually leads to valuable results.

\subsection{The matching kernel}
\label{subsec:kernmatch}

The non definite positiveness of the kernel \Kmax{} is mainly due to
the max operator.  Suard~\cite{suard-07} proposes to replace the
kernel \Kpath{} by a kernel which decreases  abruptly when the two
paths are different.  The resulting kernel is defined as:
\begin{eqnarray}
  \label{eq:2}
  \Kmatching(G_1,G_2) &=& \Kmatching(P_1,P_2) = \\ 
  && \frac{1}{\abs{P_1}} \frac{1}{\abs{P_2}}
  \sum_{h_i\in P_1} \sum_{h_j\in P_2}  \exp\left(\frac{-d^2_{path}(h_i,h_j)}{2\sigma^2}\right). \notag
\end{eqnarray}
where $d_{path}$ is the distance associated to the kernel \Kpath{} and
defined by: $d_{path}^2(h_1,h_2) = K_{path}(h_1,h_1) +
K_{path}(h_2,h_2) - 2 K_{path}(h_1,h_2)$.

The resulting function defines a definite positive kernel. This kernel
relies on the assumption that using a small value of $\sigma$, the
couple of paths with the smallest distance will predominate the others
in equation~\ref{eq:2}. This kernel may thus lead to erroneous
results if the distance are of the same order of magnitude than
$\sigma$ or if several couples of paths have nearly similar distances.

\subsection{The change detection kernel}

Desobry~\cite{Desobry2005} proposed a general approach for the
comparison of two sets which has straightforward applications in the
design of a kernel between bags (sets) of paths.  Desobry models the
two sets as the observation of two sets of random variables in a
feature space and proposes to estimate a distance between the two
distributions without explicitly building the pdf of the two sets.

The feature space considered by Desobry is based on the normalised
kernel ($K(h,h')=K_{path}(h,h')
/\sqrt{(K_{path}(h,h)K_{path}(h',h'))}$). Using such a kernel we have
$\|h\|^2_K=K(h,h)=1$ for any path. The image in the feature space of
our set of paths lies thus on an hypersphere of radius $1$ centered at
the origin (Fig.~\ref{fig:changekern}). Desobry defines a region on
this sphere by using a single class $\nu$-SVM. This region corresponds
to the density support estimate of the unknown pdf of the set of
paths~\cite{Desobry2005}. 

Using Desobry's method, two set of vectors
are thus map onto two regions of the unit sphere and the distance
between the two regions corresponds to a distance between the two
sets.  Several kernels based on this mapping have been proposed:
\begin{figure}[t]
 \mbox{  }\hfill
 \subfigure[Sets on the unit sphere]{\includegraphics[width=0.3\textwidth]{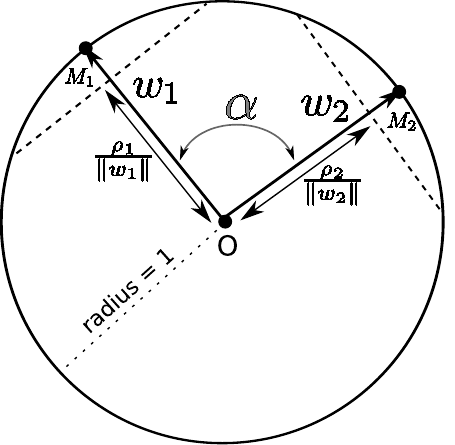}}
   \hfill
   \subfigure[original]{\includegraphics[height=0.15\textwidth]{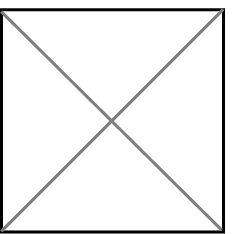}}
   \hfill
   \subfigure[edge protrusion]{\includegraphics[height=0.15\textwidth]{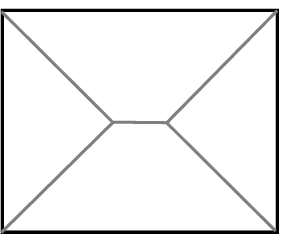}}
   \hfill
   \subfigure[node insertion]{\includegraphics[height=0.15\textwidth]{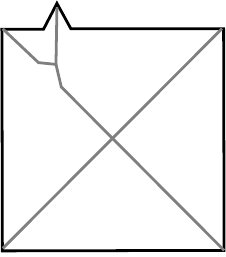}}
\hfill\mbox{ }
\caption{Separating two sets using one-class SVM (a).
  The symbols $(w_1,\rho_1)$ and $(w_2,\rho_2)$ denote the parameters
  of the two hyperplanes which are represented by dashed lines.
  Influence of small perturbations on the bag of paths ((b), (c) and
  (d))}
  \label{fig:changekern}
\end{figure}
\begin{enumerate}
\item Desobry proposed~\cite{Desobry2005} to define the distance
  between the two spherical arcs as a contrast measure defined by:
 $   d_{Desobry}^2(P_1,P_2) = \frac{\arccos\left(\frac{w_1 K_{1,2} w_2 }{\norm{w_1}\norm{w_2}}\right)}{
      \arccos\left(\frac{\rho_1}{\norm{w_1}}\right) +
      \arccos\left(\frac{\rho_2}{\norm{w_2}}\right)
    }.$
  This distance is connected to the Fisher ratio (see \cite[Section
  IV]{Desobry2005}). However, the definite positiveness of the Gaussian RBF
  kernel based on this distance remains to be shown.
\item Suard~\cite{suard-07} proposed the following kernel:
 $   K_{Suard}(G_1,G_2) = K_{Suard}(P_1,P_2) = \rho_1 \rho_2  
\sum_{h_i\in P_1} \sum_{h_j\in P_2} \alpha_{1,i}\
  K_{path}(h_i,h_j) \alpha_{2,j}$
  with $w_1 = (\alpha_{1,1},\ldots,\alpha_{1,\abs{P_1}})$ and $w_2 =(\alpha_{2,1},\ldots,\alpha_{2,\abs{P_2}})$.  

  This kernel is definite positive, but does not correspond to any
  straightforward geometric interpretation.
\end{enumerate}

\subsection{Path kernel}
All the kernels between bags of paths defined in
Section~\ref{sec:kernBag} are based on a generic kernel \Kpath{}
between paths. A kernel between two paths $h_1=(v_1,\dots,v_n)$ and
$h'=(v'_1,\dots,v'_p)$ is classically~\cite{kashima-03} built by
considering each path as a sequence of nodes and a sequence of edges.
This kernel denoted \Kclassic{} is then defined as $0$ if both paths
have not the same size and as follows otherwise:
\begin{equation}
  \label{eq:5}
  \begin{split}
    K_{classic}(h,h')=    K_v(\varphi(v_1),\varphi(v'_1)) \overset{\abs{h}}{\underset{i=2}{\prod}} K_e(\psi(e_{v_{i-1}v_i}),\psi(e_{v'_{i-1}v'_i}))
    K_v(\varphi(v_i),\varphi(v'_i))
  \end{split}
\end{equation}
where $\varphi(v)$ and $\psi(e)$ denote respectively the vectors of features
associated to the node $v$ and the edge $e$. The terms $K_v$ and
$K_e$ denote two kernels between respectively nodes and edge's
features. For the sake of simplicity, we have used Gaussian RBF
kernels between the attributes of nodes and edges
(Section~\ref{sec:Experiments}).

\section{Hierarchical kernels}
\label{sec:kernhier}
Since the main focus of this paper is a new kernel method for shape
classification, the construction of skeletal graphs from shapes has
been adressed using classical methods. We first build a skeleton using
the method proposed by Siddiqi~\cite{sidiqqi-99}. However the graph we
build from the skeleton does not correspond to the shock graph
proposed by Siddiqi. Indeed, this graph provides a precise description
of the shape but remains sensitive to small perturbations of the
boundary.  We rather use the construction scheme proposed by
Suard~\cite{suard-07} and Ruberto~\cite{ruberto-04} which
consists to select as node all the pixels of the skeleton which
correspond to end points or junctions.  These nodes are then connected
by edges, each edge being associated to one branch of the skeleton.
Given a skeletal graph $G$ we valuate each of its edge by an additive
weight measure and we consider the maximal spanning tree $T$ of $G$.
The bag of path associated to $G$ is built on the tree $T$. Note that,
the skeletonization being homotopic we have $G=T$ if the $2D$ shape
does not contain any hole.

\subsection{Bag of path kernel}
\label{sec:kernPath}
None of the bag of path kernels proposed by Desobry or Suard
(Section~\ref{sec:kernBag}) is both definite positive and provides a
clear geometrical interpretation. We thus propose a new kernel based
on the following distance:
\begin{equation}
  \label{eq:3}
  d_{change}^2(P_1,P_2) = \arccos\left( \frac{w_1 K_{1,2} w_2 }{\norm{w_1}\norm{w_2}}\right).
\end{equation}
This distance corresponds to the angle $\alpha$ between the two mean
vectors $w_1$ and $w_2$ of each region (Fig.~\ref{fig:changekern}).
Such an angle may be interpreted as the geodesic distance between two
points on the sphere and has thus a clear geometrical interpretation.
Based on this distance we use the Gaussian RBF kernel:
\begin{equation}
  \label{eq:4}
  K_{change}(G_1,G_2) = K_{change}(P_1,P_2) = \exp\left( \frac{-d^2_{change}(P_1,P_2)}{2\sigma^2} \right).
\end{equation}
This kernel is definite positive since the normalized scalar product
is positive definite and $\arccos$ is bijective on $[0,1]$. The
Gaussian RBF kernel based on this distance is thus definite positive
(see~\cite{Berg1984} for further details).

\subsection{Hierarchical kernel between paths}
\label{subsec:kerhier}

A mentioned in Section~\ref{sec:intro}, the use of kernels between
bags of paths within the shape matching framework relies on the
assumption that the graphs associated to two similar shapes share a
large amount of similar paths. This assumption is partially false
since a small amount of structural noise may have important
consequences on the set of paths. Let us for example, consider the
small deformation of the square (Fig.~\ref{fig:changekern}(b))
represented on Fig.~\ref{fig:changekern}(c). This small deformation
transforms the central node in Fig.~\ref{fig:changekern}(b) into an
edge (Fig.~\ref{fig:changekern}(c)).  Consequently graphs associated
to these two shapes only share two paths of length $2$ (the ones which
connect the two corners on the left and right sides).  In the same
way, a small perturbation of the boundary of the shape may add
branches to the skeleton(Fig.~\ref{fig:changekern}(d)). Such
additional branches i) split existing edges into two sub edges by
adding a node and ii) increase the size of the bag of path either by
adding new paths or by adding edges within existing paths.

The influence of small perturbations of the shape onto an existing set
of paths may thus be modeled by node and edge insertions along these
paths. In order to get a path kernel robust against structural noise
we associate to each path a sequence of successively reduced paths,
thus forming a hierarchy of paths. Our implicit assumption is that, if
a path has been elongated by structural noise one of its reduced
version should corresponds to the original path.

The reduction of a path is performed either by node removal or edge
contraction along the path.  Such a set of reduction operations is
compatible with the taxinomy of topological transition of the skeleton
compiled by Giblin and Kimia~\cite{giblin-99}. Note that, since all
vertices have a degree lower than $2$ along the path these operations
are well defined. In order to select both the type of operation and
the node or the edge to respectively remove or contract we have to
associate a weight to each node and edge which reflects its importance
according to the considered path and the whole graph.

\begin{figure}[t]
  \centering
  \includegraphics[width=0.7\textwidth]{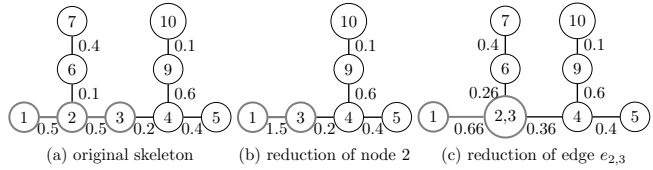}
  \caption{Different reductions of a path (in gray) within a skeletal tree.}
  \label{fig:path1}
\end{figure}

Let us consider a skeletal graph $G$, its associated maximal spanning
tree $T$ and a path $h=(v_1,\dots,v_n)$ within $T$. We valuate each
operation on $h$ as follows:
\begin{description}
\item[Node removal :] Let us denote by $v_i$, $i\in \{2,\dots,n-1\}$
  the removed node of the path $h$. The node $v_i$ has a degree
  greater than $2$ in $T$ by construction. Our basic idea consists to
  valuate the importance of $v_i$ by the total weight of the
  additional branches which justify its existence within the path $h$.
  For each neighbor $v$ of $v_i$ not equal to $v_{i-1}$ nor $v_{i+1}$
  we compute the weight $W(v)$ defined as the addition of the weight
  of the tree rooted on $v$ in $T-\{e_{v_iv}\}$ and the weight of $e_{v_iv}$.
  This tree is unique since $T$ is a tree. The
  weight of the node $v_i$ (and the cost of its removal) is then
  defined as the sum of weight $W(v)$ for all neighbors $v$ of $v_i$
  (excluding $v_{i-1}$ and $v_i$).

  After the removal of this node the edges $e_{v_{i-1}v_i}$ and
  $e_{v_iv_{i+1}}$ are concatenated into a single edge in the new path
  $h'$. The weight of this new edge is defined as the sum of the
  weight of the edges $e_{v_{i-1}v_i}$, $e_{v_iv_{i+1}}$ and the weight
  of the node $v_i$ (Fig.~\ref{fig:path1}(a) and (b)).

\item[Edge contraction :] The cost of an edge contraction is measured
  by the relevance of the edge which is encoded by its weight.  Let us
  denote by $e_{v_iv_{i+1}}$, $i<n$ the contracted edge of the path
  $h=(v_1,\dots,v_n)$. In order to preserve the total weight of the
  tree after the contraction, the weight of the edge $e_{v_iv_{i+1}}$
  is equally distributed among the edges of $T$ incident to $v_i$ and
  $v_{i+1}$:
  \[
  \forall e \in \iota(v_i)\cup\iota(v_{i+1})-\{e_{v_iv_{i+1}}\} \quad
  w'(e)=w(e)+\frac{w(e_{v_iv_{i+1}})}{d(v_i)+d(v_{i+1})-2}
  \]
  where $\iota(v)$ and $d(v)$ denote respectively the set of edges
  incident to $v$ and the cardinal of this set (the vertex's degree).
  The symbol $w(e)$ denotes the weight of the edge $e$.

  For example, the contraction of the edge $e_{2,3}$ in
  Fig.~\ref{fig:path1}(a) corresponds to a cost of $w(e_{2,3})=.5$.
  The contraction of this edge induces the incrementation of the
  edge's weights $w(e_{2,1})$,$w(e_{2,6})$,$w(e_{3,4})$ by
  $.5/3\approx.16$
\end{description}

Any additive measure encoding the relevance of a branch of the
skeleton may be used as a weight. We choose to use the measure defined
by Torsello~\cite{Torsello2004} which associates to each branch of the
skeleton (an thus to each edge) the length of the boundaries which
contributed to the creation of this branch. Such a measure initially
defined for each pixel of the skeleton is trivially additive.

Let us denote by $\kappa$ the function which applies the cheapest
operation on a path. The successive applications of the function
$\kappa$ associate to each path $h$ a sequence of reduced paths
$(h,\kappa(h),\dots,\kappa^D(h))$ where $D$ denotes the maximal number
of reductions.  Using \Kclassic{} for the path comparison, we define
the kernel \Kedit{} as the mean value of kernels between reduced paths
of equal length. Given two paths $h$ and $h'$, this kernel is thus
equal to $0$ if $| |h| - |h'| | >D$. Indeed, in this case the maximal
reduction of the longuest path remains longuer than the shortest one.
Otherwise, $| |h| - |h'|| \leq D$, and $K_{edit}(h,h')$ is defined as:
\begin{equation}
  \label{eq:9}
  K_{edit}(h,h') = 
  \frac{1}{D+1} \sum_{k=0}^{D} \sum_{l=0}^{D}
  K_{classic}(\kappa^k(h),\kappa^l(h'))
\end{equation} 

This kernel is proportional (by a factor $D+1$) to a sum of
$R$-convolution kernels \cite[Lemma 1]{haussler-99} and is thus
definite positive.

Since \Kclassic{} is equal to $0$ for paths of different lengths,
\Kedit{} is indeed equal to a sum of kernels between reduced paths of
equal length. For example, given two paths $h$ and $h'$ whose
respective length is equal to $4$ and $3$ we have for $D=2$:
\[
\Kedit(h,h')=\frac{1}{3}\left[\Kclassic(\kappa(h),h')+\Kclassic(\kappa^2(h),\kappa(h'))\right]
\]

\section{Experiments}
\label{sec:Experiments}

We used the following features for our experiments: Each node is
weighted by its distance to the gravity center of the shape and each
edge is assocated to a vector of two features: The first feature
corresponds to the edge's weight (section~\ref{subsec:kerhier}). The
second feature is the angle between the straight line passing through
the two nodes of the edge and the principal axis of the shape. These
experiments are based on the LEMS~\cite{LEMS} database which consists
of 99 objects divided into 9 classes.

We defined three kernels for these experiments: The kernel \KmaxC{}
based on a conjoint use of the kernels \Kmax{} (equation~\ref{eq:6})
and \Kclassic{} (equation~\ref{eq:5}) has been introduced by
Suard~\cite{suard-07}. The kernel \KchangeC{} based on a conjoint use
of the kernels \Kchange{} (equation~\ref{eq:4}) and \Kclassic{} allows
to evaluate the performances of the kernel \Kchange{} compared to the
kernel \Kmax{}. Finally, the kernel \Knew{} is based on a conjoint
used of the two kernels \Kedit{} and \Kchange{} proposed in this
paper. The kernel \Kclassic{} is defined by the two parameters
$\sigma_{edge}$ and $\sigma_{vertex}$ respectively used by the
Gaussian RBF kernels on edges and vertices.  The kernel \KmaxC{} does
not require additional parameters while \KchangeC{} is based on a
$\nu$-SVM and requires thus the parameter $\nu$. It additionally
requires the parameter $\sigma_{change}^{classic}$ used by the RBF
kernel in~ equation~\ref{eq:4}.  The kernel \Kedit{} requires the two
parameters $\sigma_{edge}$ and $\sigma_{vertex}$ used by $\Kclassic$
together with the maximal number of edition ($D$). Finally, the kernel
\Knew{} requires as \Kchange{} the two additional parameters $\nu$ and
$\sigma_{change}^{new}$(equation~\ref{eq:4}).  These parameters have
been fixed to the following values in the experiments described below:
$D=2$, $\sigma_{edge} = \sigma_{vertex} = 0.1$, $\nu = 0.9$,
$\sigma_{change}^{new} = 0.3$ and $\sigma_{change}^{classic} = 1.0$.
The parameters $\sigma_{edge}$ and $\sigma_{vertex}$ are common to all
kernels. The remaining parameters have been been set in order to
maximize the performances of each kernel on the experiments below.

\begin{figure}[t]
  \centering
  \subfigure[hands,tools,dudes]{
    \begin{tabular}[c]{ccccc}
      \includegraphics[width=0.04\textwidth]{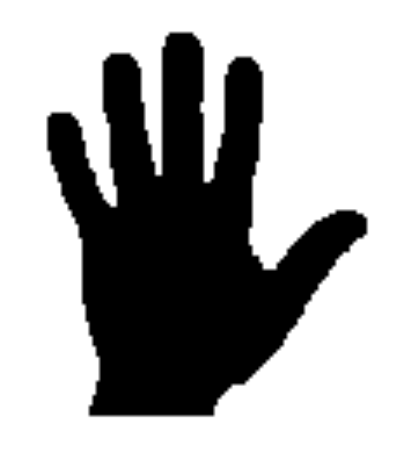} &
      \includegraphics[width=0.04\textwidth]{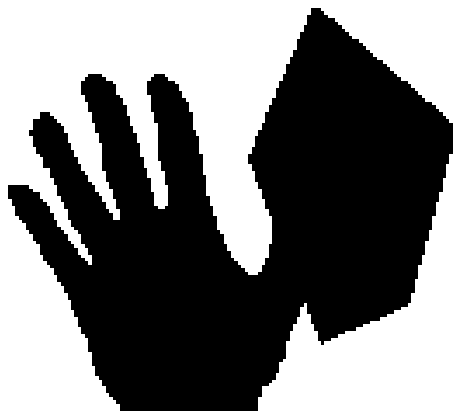}&
      \includegraphics[width=0.04\textwidth]{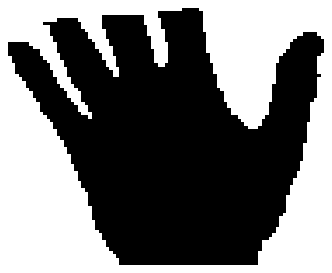} &
      \includegraphics[width=0.04\textwidth]{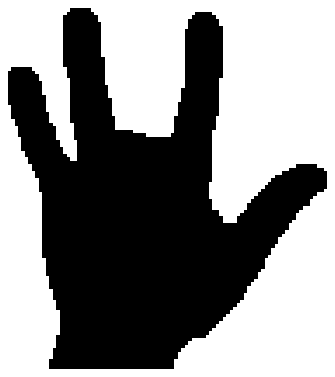} &
      \includegraphics[width=0.04\textwidth]{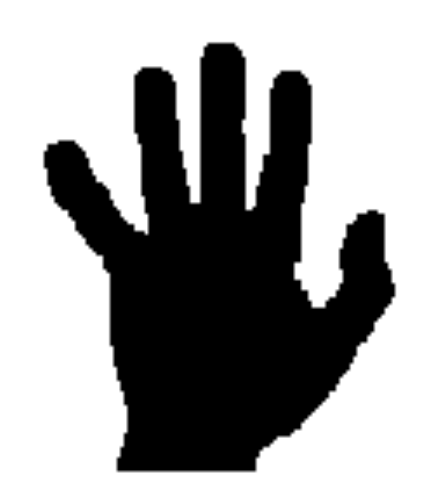} \\
      \includegraphics[width=0.04\textwidth]{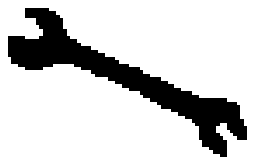} &
      \includegraphics[width=0.04\textwidth]{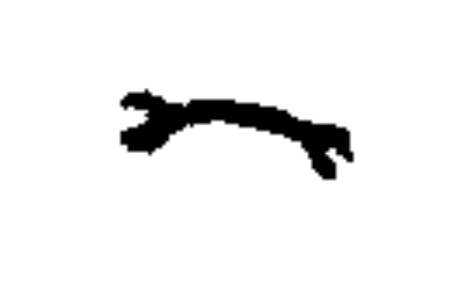} &
      \includegraphics[width=0.04\textwidth]{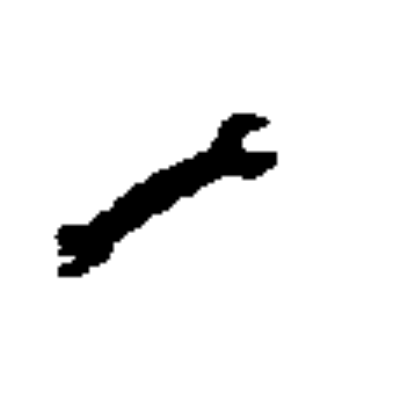} &
      \includegraphics[width=0.04\textwidth]{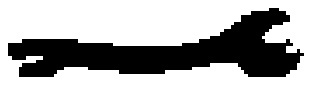} &
      \includegraphics[width=0.04\textwidth]{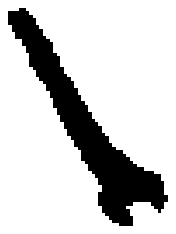} \\
      \includegraphics[width=0.04\textwidth]{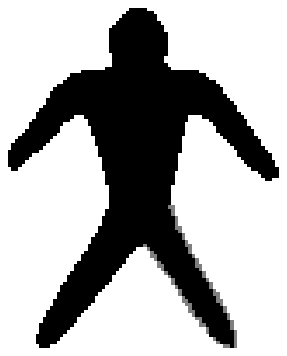} &
      \includegraphics[width=0.04\textwidth]{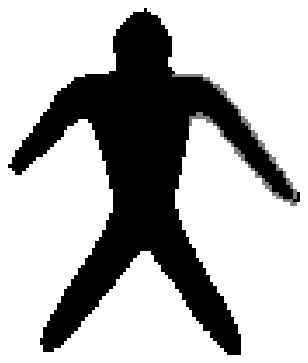} &
      \includegraphics[width=0.04\textwidth]{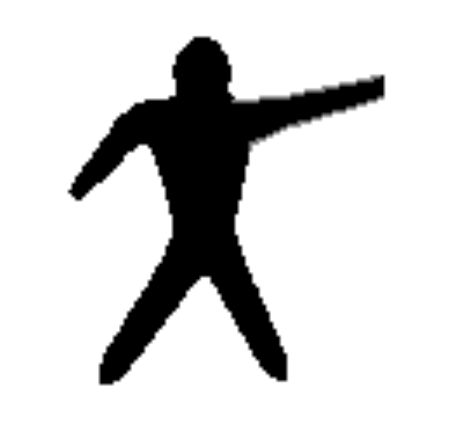} &
      \includegraphics[width=0.04\textwidth]{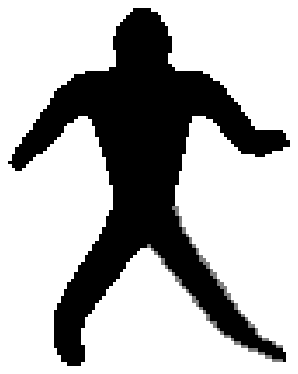} &
      \includegraphics[width=0.04\textwidth]{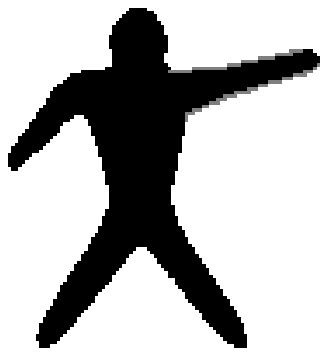} \\
    \end{tabular}
  } \hfill \subfigure[sorted distances to the hand]{
    \begin{tabular}{cll}
      (1)&
  \begin{tabular}[ht]{c|cccccccccc}
    \includegraphics[width=0.04\textwidth]{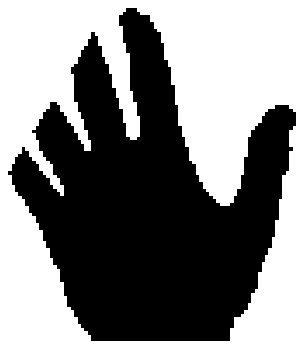} &
    \includegraphics[width=0.04\textwidth]{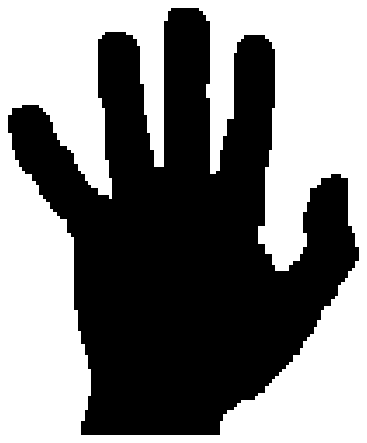} &
    \includegraphics[width=0.04\textwidth]{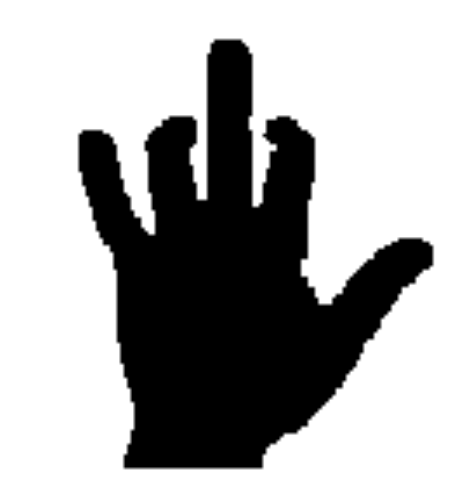} &
    \includegraphics[width=0.04\textwidth]{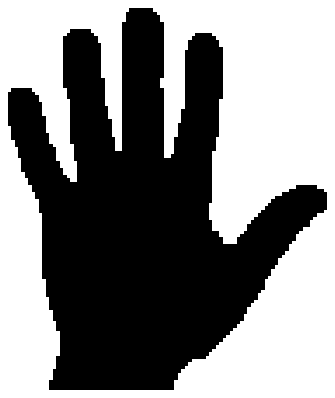} &
    \includegraphics[width=0.04\textwidth]{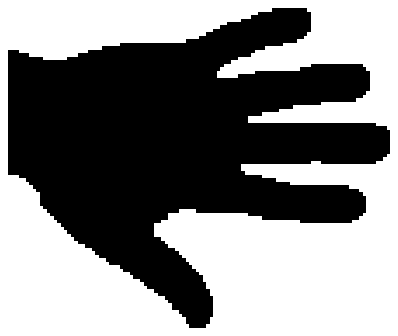} &
    \includegraphics[width=0.04\textwidth]{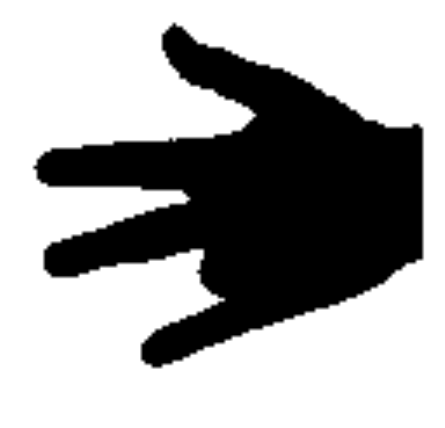} &
    \includegraphics[width=0.04\textwidth]{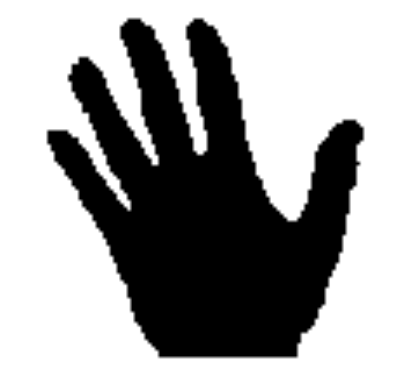} &
    \includegraphics[width=0.04\textwidth]{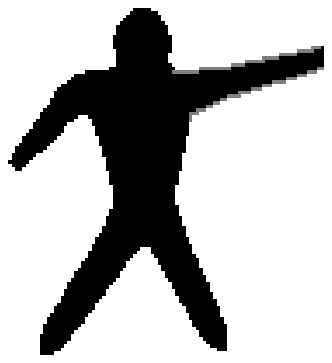} &
    \includegraphics[width=0.04\textwidth]{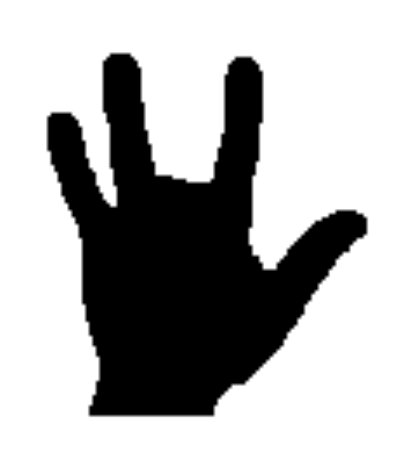} &
    \includegraphics[width=0.04\textwidth]{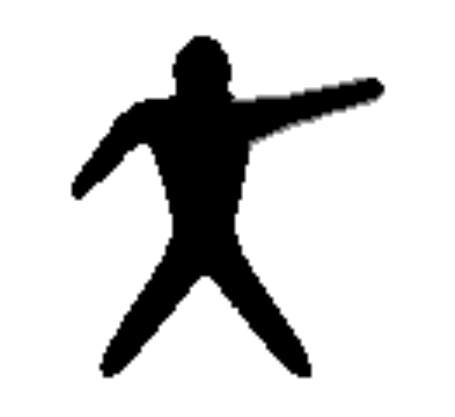} &
    \includegraphics[width=0.04\textwidth]{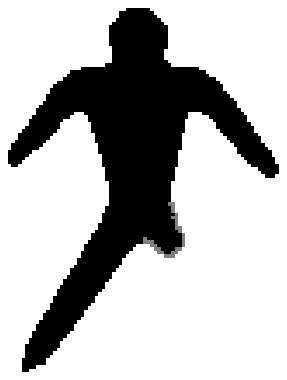} \\
  \end{tabular} &$\eta=7$\\
  (2)&
  \begin{tabular}[ht]{c|cccccccccc}
    \includegraphics[width=0.04\textwidth]{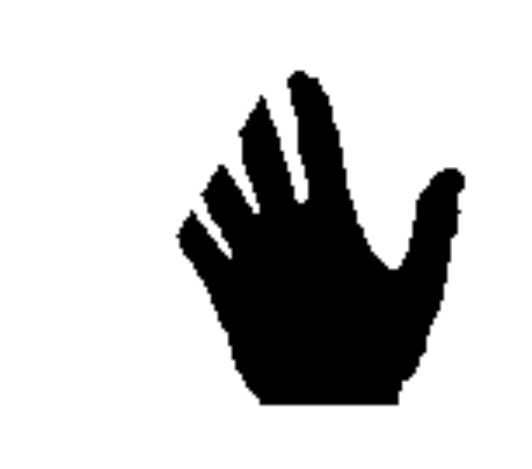} &
    \includegraphics[width=0.04\textwidth]{handbent1} &
    \includegraphics[width=0.04\textwidth]{hand} &
    \includegraphics[width=0.04\textwidth]{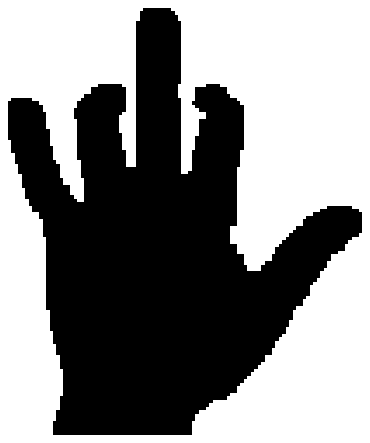} &
    \includegraphics[width=0.04\textwidth]{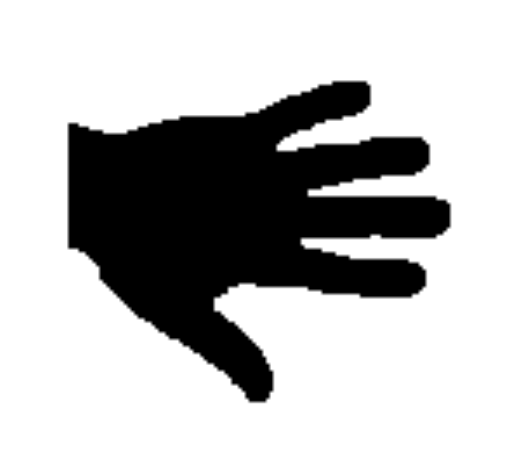} &
    \includegraphics[width=0.04\textwidth]{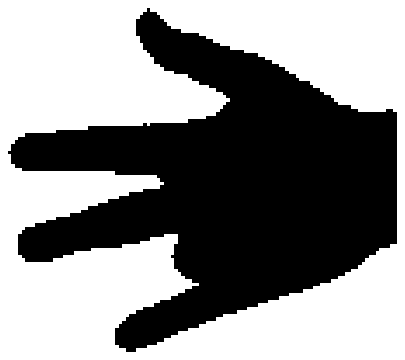} &
    \includegraphics[width=0.04\textwidth]{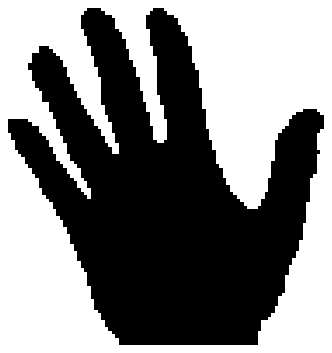} &
    \includegraphics[width=0.04\textwidth]{dude2} &
    \includegraphics[width=0.04\textwidth]{handdeform} &
    \includegraphics[width=0.04\textwidth]{dude4} &
    \includegraphics[width=0.04\textwidth]{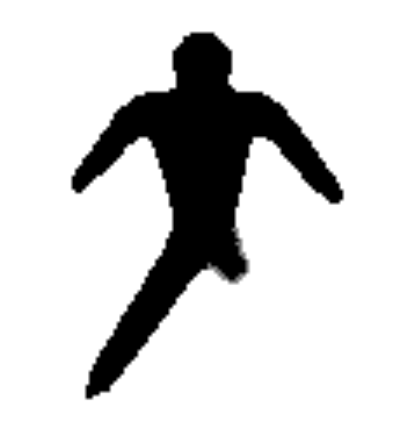} \\
  \end{tabular}&$\eta=7$\\
  (3)&
  \begin{tabular}[ht]{c|cccccccccc}
    \includegraphics[width=0.04\textwidth]{hand2occ3} &
    \includegraphics[width=0.04\textwidth]{handbent1} &
    \includegraphics[width=0.04\textwidth]{hand} &
    \includegraphics[width=0.04\textwidth]{handbent2} &
    \includegraphics[width=0.04\textwidth]{hand90} &
    \includegraphics[width=0.04\textwidth]{hand2} &
    \includegraphics[width=0.04\textwidth]{handdeform2} &
    \includegraphics[width=0.04\textwidth]{handdeform} &
    \includegraphics[width=0.04\textwidth]{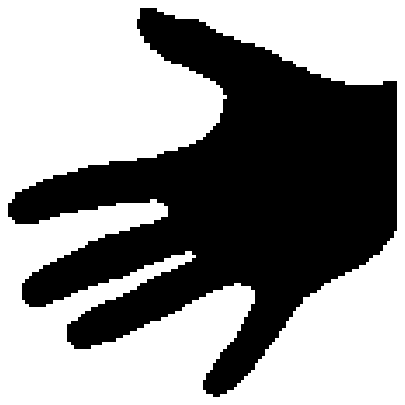} &
    \includegraphics[width=0.04\textwidth]{dude2} &
    \includegraphics[width=0.04\textwidth]{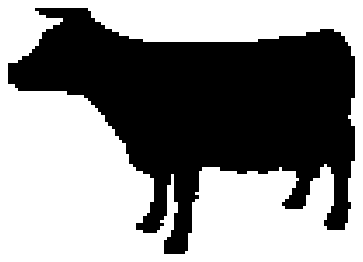} \\
  \end{tabular}&$\eta=9$
\end{tabular}
}
\caption{Five representative shapes of the classes hands, tools and
  dudes of the LEMS database (a), and (b) the 10 closest shapes from an
  hand using the distances induced by the kernels \KmaxC{} (1),
  \KchangeC{} (2) and \Knew{} (3).}
  \label{fig:hand}
\end{figure}

Our first experiment compares the distance induced by each kernel $k$ and defined as
$d^2(x,x') = k(x,x) + k(x',x') - 2k(x,x')$. The mean number of matches for each class is
defined as follows: For each shape of the selected class we sort all the shapes of the
database according to their distances to the selected shape using an ascending order.  The
number of good matches of the input shape is then defined as the number of shapes ranked
before the first shape which belongs to a different class than the selected one.  For
example, the $10$ nearest neighbors of a hand sorted in an ascending order are represented
in Fig.~\ref{fig:hand}(b), the number of good matches of each shape is indicated on the
right of the figure.  Note that the greater number of good match being obtained for the
kernel \Knew{}. The mean number of good matches of a class is defined as the mean value of
the number of good matches for each shape of the class. The different values represented
in Tab.~\ref{tab:match}(a) represent the mean values of these number of good matches for
the classes: hands, tools and dudes (Fig.~\ref{fig:hand}(a)). As indicated by
Tab.~\ref{tab:match}(a), the kernel \KmaxC{} provides stable results but is sensitive to
slight perturbations of the shapes as the ones of the class dudes and cannot handle the
severe modifications of the hands. The kernel \KchangeC{} leads to roughly similar results
on the different classes.  Though not presented here, the kernel \KmatchingC{}
(equation~\ref{eq:2}) gives worst results than the others kernels. This result may be
explained by the drawbacks of this kernel (Section~\ref{subsec:kernmatch}). The kernel
\Knew{} always provides the best results with a good robustness to perturbation on dudes
and hands.

\begin{table}[b]
  \centering
  \mbox{ }\hfill
  \subtable[Mean number of good matches.]{
    \begin{minipage}[b]{.33\textwidth}  
      \begin{tabular}{l|c|c|c}
        & Hands & Tools & Dudes \\
        \hline $\KmaxC$ & 4.81 & 6.18 & 6.36 \\
        \hline $\KchangeC$ & 5.27 & 5.45 & 6.36 \\
        \hline $\Knew$ & 7.09 & 9.82 & 6.36 \\
        \hline
      \end{tabular}
    \end{minipage}
  }  \hfill
  \subtable[Number of recognized shapes in one class.]{
    \begin{minipage}[c]{0.4\textwidth}
      \begin{tabular}{l|c|c|c}
        & Hands & Tools & Dudes \\
        \hline $\KmaxC$ & 7 & 11 & 10 \\
        \hline $\KchangeC$ & 7 & 10 & 10 \\
        \hline $\Knew$ & 9 & 11 & 11 \\
        \hline
      \end{tabular}
      \end{minipage}
  }
  \hfill\mbox{ }
  \caption{Kernels evaluation based on distance (a) and classification (b) criteria.}
  \label{tab:match}
\end{table}

Our second experiment evaluates performances of each kernel within a classification
framework. To this end, we have trained a SVM on 5 shapes of each of the three classes: dudes,
hands, tools on one side and one model of each of the 6 remaining classes on
the other side. The SVM margin parameter was selected in order to maximize the number of
true positive while having no false positive. Tab.~\ref{tab:match}(b) shows the number of well
classified shapes for each class. The kernel \Knew{} gives the best performances
especially for the hands where the two missing shapes are the more perturbed ones. The two
others kernels present good results and are competitive when shapes are not strongly deformed.
This experiment confirms the robustness of our kernel against perturbed shapes.

\section{Conclusion}

The bag of path approach is based on a decomposition of the complex
graph structure into a set of linear objects (paths). Such an approach
benefits of recent advances in both string and vectors kernels.  Our
graph kernel based on a hierarchy of paths is more stable to small
perturbations of the shapes than kernels based solely on a bag of
paths. Our notion of path's hierarchy is related to the graph edit
distance through the successive rewritings of a path. Our kernel is
thus related to the ones introduced by Neuhaus and Bunke. 

\bibliographystyle{splncs}
\bibliography{paperbib}

\begin{thebibliography}{10}

\bibitem{sharvit-98}
Sharvit, D., Chan, J., Tek, H., Kimia, B.:
\newblock Symmetry-based indexing of image databases.
\newblock Journal of Visual Communication and Image Representation
  \textbf{9}(4) (Dec. 1998)  366--380

\bibitem{sidiqqi-99}
Siddiqi, K., Shokoufandeh, A., Dickinson, S.J., Zucker, S.W.:
\newblock Shock graphs and shape matching.
\newblock Int. J. Comput. Vision \textbf{35}(1) (1999)  13--32

\bibitem{pelillo-99}
Pelillo, M., Siddiqi, K., Zucker, S.:
\newblock Matching hierarchical structures using association graphs.
\newblock IEEE Trans. on PAMI \textbf{21}(11) (Nov 1999)  1105--1120

\bibitem{bunke-07}
Bunke, H., Riesen, K.:
\newblock A family of novel graph kernels for structural pattern recognition.
\newblock In: CIARP. (2007)  20--31

\bibitem{Neuhaus-06}
Neuhaus, M., Bunke, H.:
\newblock Edit distance based kernel functions for structural pattern
  classification.
\newblock Pattern Recognition \textbf{39} (2006)  1852--1863

\bibitem{suard-07}
Suard, F., Rakotomamonjy, A., Bensrhair, A.:
\newblock Kernel on bag of paths for measuring similarity of shapes.
\newblock In: European Symposium on Artificial Neural Networks, Bruges-Belgique
  (April 2007)

\bibitem{Haasdonk2005}
Haasdonk, B.:
\newblock Feature space interpretation of svms with indefinite kernels.
\newblock IEEE PAMI \textbf{27}(4) (April 2005)  482--492

\bibitem{Desobry2005}
Desobry, F., Davy, M., Doncarli, C.:
\newblock An online kernel change detection algorithm.
\newblock IEEE TSP \textbf{53}(8) (August 2005)  2961--2974

\bibitem{kashima-03}
Kashima, H., Tsuda, K., Inokuchi, A.:
\newblock Marginalized kernel between labeled graphs.
\newblock In: In Proc. of the Twentieth International conference on machine
  Learning. (2003)

\bibitem{ruberto-04}
Ruberto, C.D.:
\newblock Recognition of shapes by attributed skeletal graphs.
\newblock Pattern Recognition \textbf{37}(1) (2004)  21--31

\bibitem{Berg1984}
Berg, C., Christensen, J.P.R., Ressel, P.:
\newblock Harmonic Analysis on Semigroups.
\newblock Springer-Verlag (1984)

\bibitem{giblin-99}
Giblin, P.J., Kimia, B.B.:
\newblock On the local form and transitions of symmetry sets, medial axes, and
  shocks.
\newblock In: Seventh Internat. Conf. on Computer Vision. (1999)  385--391

\bibitem{Torsello2004}
Torsello, A., Handcock, E.R.:
\newblock A skeletal measure of 2d shape similarity.
\newblock CVIU \textbf{95} (2004)  1--29

\bibitem{haussler-99}
Haussler, D.:
\newblock Convolution kernels on discrete structures.
\newblock Technical report, Department of Computer Science, University of
  California at Santa Cruz (1999)

\bibitem{LEMS}
LEMS:
\newblock shapes databases.
\newblock http://www.lems.brown.edu/vision/software/index.html

\end{thebibliography}

\end{document}